# A NEW THREE-DOF PARALLEL MECHANISM: MILLING MACHINE APPLICATIONS

D. CHABLAT and P. WENGER

*Institut de Recherche en Communication et Cybernétique de Nantes*
*1, rue la Noë 44072 NANTES - FRANCE*
*Phone +33 240376947 - Fax +33 240376930*
`Email : Damien.Chablat@ircyn.ec-nantes.fr`

**Abstract** - This paper describes a new parallel kinematic architecture for machining applications, namely, the *orthoglide*. This machine features three fixed parallel linear joints which are mounted orthogonally and a mobile platform which moves in the Cartesian x-y-z space with fixed orientation. The main interest of the orthoglide is that it takes benefit from the advantages of the popular PPP serial machines (regular Cartesian workspace shape and uniform performances) as well as from the parallel kinematic arrangement of the links (less inertia and better dynamic performances), which makes the orthoglide well suited to high-speed machining applications. Possible extension of the orthoglide to 5-axis machining is also investigated.

## 1    Introduction

Parallel kinematic machines (PKM) are commonly claimed to offer several advantages over their serial counterpart, like high structural rigidity, high dynamic capacities and high accuracy. On the other hand, they generally suffer from a reduced operational workspace due to the presence of internal singularities or self-collisions.

PKM were first introduced by Gough in 1956 and have been used for many years as flight simulators, because of their higher load carrying capacity (Stewart 1965), (Gosselin and Wang 1998). Parallel robotic manipulators appeared later with the first industrial application proposed by Clavel (Clavel 1988). Parallel kinematic machine tools attract the interest of more and more researchers and companies. Since the first prototype presented in 1994 during the IMTS in Chicago by Gidding&Lewis (the Variax), many other prototypes have appeared as could be seen during the last World Exhibition EMO'99 which was held in Paris in May 1999. A recent comparative study, conducted on the basis of simple planar mechanisms, shows that certain parallel kinematic structures do have potential advantages over their serial counterparts (Wenger et al 1999). Despite this, it is worth noting that many users of machine tools are still not convinced by the potential benefits of PKM. Most industrial 3-axis machine tools have a PPP kinematic architecture with orthogonal joint axes. Thus, the motion of the tool in the space is quite simply related to the motion of the three actuated axes. Also, the performances (e.g. maximum speeds, forces, accuracy and rigidity) are constant in the most part of the Cartesian workspace, which is a parallelepiped. In contrast, the common feature of most existing PKM is that the Cartesian workspace shape is of complex geometry and the motion of the tool and the motion of the actuated axes are not simply connected. More precisely, the Jacobian matrix which relates the joint rates and the output velocities is not constant and not isotropic. Consequently, the performances may vary considerably for different points in the Cartesian workspace and for different directions at one given point, which is a serious drawback for machining applications. The orthoglide studied in this paper is designed in order keep the regularity of the Cartesian workspace shape as well as the uniformity of performances of the PPP machine tools, while taking benefit from the parallel kinematic arrangement of the links.

The organisation of this paper is as follows. Next section is devoted to the presentation of existing PKM and of the orthoglide. Section 3 investigates kinematic performances of the orthoglide, especially manipulability and Cartesian workspace. Possible extensions to 5-axis PKM of the orthoglide are discussed in section 4. Last section concludes this paper.

## 2    Presentation of the orthoglide

### 2.1    Existing PKM

There are many possible types of PKM architectures which find applications in motion simulators, robotic manipulators and more recently in machine tools (Merlet 1997). In the context of machine tool applications, most existing prototypes or commercial PKM can be classified into two general families: (i) PKM with fixed foot points and variable strut lengths and (ii) PKM with fixed length struts and moveable foot points.

The first family comprises the so-called hexapod machines which, in fact, feature a Gough-Stewart platform architecture. Numerous examples of hexapods PKM exist: the VARIAX-Hexacenter (Gidding&Lewis), the CMW300 (Compagnie Mécanique des Vosges), the TORNADO 2000 (Hexel), the MIKROMAT 6X (Mikromat/IWU), the hexapod OKUMA (Okuma), the hexapod G500 (GEODETIC). In this first family, we find also hybrid architectures with a 2-axis wrist mounted in series with a 3-DOF parallel structure (e.g the TRICEPT 805, Neos Robotics).

The second family (ii) of PKM has been more recently investigated. The most famous PKM of this category is the HEXAGLIDE (ETH Zürich) which features six parallel (also in the geometrical sense) and coplanar linear joints. The HexaM (Toyota) is another example with non coplanar linear joints. A 3-axis translational version of the hexaglide is the TRIGLIDE (Mikron), which has three coplanar and parallel linear joints. Another 3-axis translational PKM is proposed by the ISW Uni Stuttgart with the LINAPOD. This PKM has with three vertical (non coplanar) linear joints. The URANE SX (Renault Automation) and the QUICKSTEP (Krause & Mauser) are 3-axis PKM with three non coplanar horizontal linear joints. A hybrid parallel/serial PKM with three parallel inclined linear joints and a two-axis wrist is the GEORGE V (IFW Uni Hannover).

To be complete, one should add the ECLIPSE machining center which does not fall in the aforementioned two PKM families. This is a 6-DOF over actuated machine with three vertical struts which can move independently on an horizontal circular prismatic joint.

More detailed information about all these PKM can be found at http://www.ifw.uni-hannover.de/robotool, http://www.renault-automation.com, http://www.krause-mauser.com.

### 2.2    The orthoglide

The orthoglide presented in this paper belongs to the family of 3-axis translational PKM with variable foot points and fixed length struts (figure 1). This machine has three identical legs which are PRRRR chains (figure 2). The actuated joints are the three orthogonal linear joints. These joints can be actuated by means of linear motors or by conventional rotary motors with ball screws. The output body is connected to the prismatic joints through a set of three parallelograms, so that it can moves only in translation (note that two parallelograms would be enough). An important feature of this PKM is the symmetry of the design (the three chains are identical, in particular, the lengths $B_iC_i$ are equal) and the simplicity of the kinematic chains (all joints are simple one-DOF joints), which should contribute to lower the manufacturing cost of such a design. Also, the orthoglide is free of singularities and self-collisions.

The design process which led us to this kinematic structure is a similar to the one we applied for a 2-axis machine (Chablat et al 2000). The main idea is to produce a parallel kinematic machine which be as close as possible to a serial PPP machine. Serial PPP machines are nice since their kinematics is quite simple and the displacements of the tool are intuitive. Furthermore, the Cartesian workspace is very simple since it is a parallelepiped defined by the limits of the actuated joints. Another interesting feature of PPP machines is the uniformity of the performances over the Cartesian workspace. However, the main drawback is due to the serial kinematic arrangement of the links: one link has to support and move the following link in the chain, which increases the total moving masses, and thus limit the dynamic performances of such machines. In the context of rapid machining, the parallel kinematic layout of the links is an interesting feature. The orthoglide has three orthogonal linear joints like conventional PPP machines but they have been put in parallel. The design is such that the Jacobian matrix which relates the joint rates and the Cartesian velocities is isotropic at the center point of the Cartesian workspace. At this point, the orthoglide is kinematically equivalent to a serial PPP machine. Furthermore, the design has been optimised such that, in the rest of the Cartesian workspace, the conditioning of the aforementioned Jacobian matrix remains under a reasonable limit. In particular, the singularities are sufficiently far away from the Cartesian workspace limits.

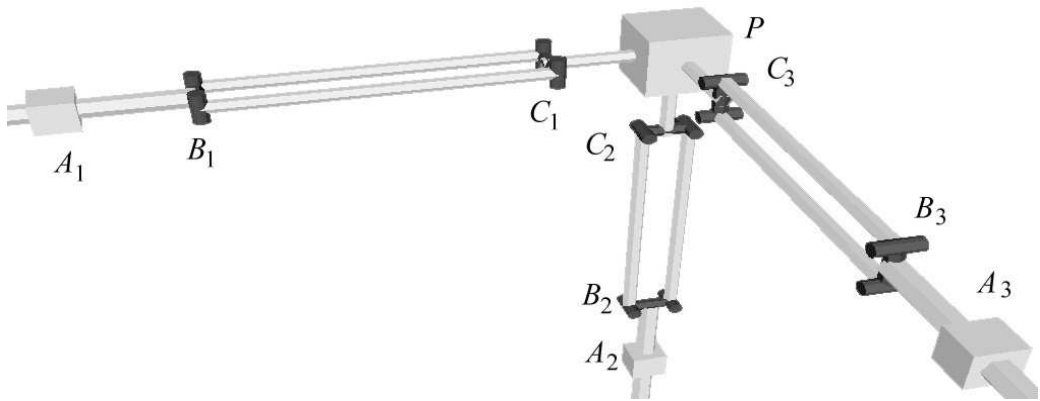

*Figure 1: orthoglide manipulator*

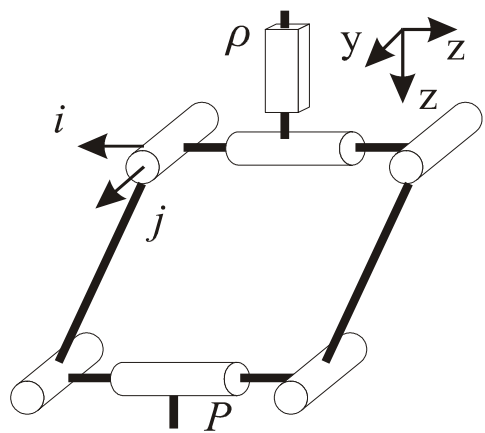

*Figure 2: one leg of the orthoglide mechanism*

### 2.3    Kinematic equations and singularity analysis

Let $\dot{\mathbf{p}}$ be defined as the vector of actuated joint rates and $\dot{\mathbf{p}}$ as the vector of velocity of point $P$:

$$\dot{\mathbf{p}} = \begin{bmatrix} \dot{\rho}_1 \\ \dot{\rho}_2 \\ \dot{\rho}_3 \end{bmatrix} \text{ and } \dot{\mathbf{p}} = \begin{bmatrix} \dot{x} \\ \dot{y} \\ \dot{z} \end{bmatrix}$$

The velocity $\dot{\mathbf{p}}$ of the point $P$ can be written in three different ways. By traversing the closed-loop $A_iB_iC_iP$ in the three possible directions, we obtain:

$$\dot{\mathbf{p}} = \frac{\mathbf{b}_1 - \mathbf{a}_1}{\|\mathbf{b}_1 - \mathbf{a}_1\|} \dot{\rho}_1 + (\dot{\theta}_1 \, \mathbf{i}_1 + \dot{\beta}_1 \, \mathbf{j}_1) \times (\mathbf{c}_1 - \mathbf{b}_1) \tag{1a}$$

$$\dot{\mathbf{p}} = \frac{\mathbf{b}_2 - \mathbf{a}_2}{\|\mathbf{b}_2 - \mathbf{a}_2\|} \dot{\rho}_2 + (\dot{\theta}_2 \, \mathbf{i}_2 + \dot{\beta}_2 \, \mathbf{j}_2) \times (\mathbf{c}_2 - \mathbf{b}_2) \tag{1b}$$

$$\dot{\mathbf{p}} = \frac{\mathbf{b}_3 - \mathbf{a}_3}{\|\mathbf{b}_3 - \mathbf{a}_3\|} \dot{\rho}_3 + (\dot{\theta}_3 \, \mathbf{i}_3 + \dot{\beta}_3 \, \mathbf{j}_3) \times (\mathbf{c}_3 - \mathbf{b}_3) \tag{1c}$$

where $\mathbf{a}_i$, $\mathbf{b}_i$ and $\mathbf{c}_i$ represent the position vector of the points $A_i$, $B_i$ and $C_i$, respectively, for $i$=1, 2, 3 ($A_i$ and $B_i$ cannot coincide).

We want to eliminate the two idle joint rates $\dot{\theta}_i$ and $\dot{\beta}_i$ from equations (1a), (1b) and (1c), which we do upon dot-multiplying eq. (1i) by $\mathbf{c}_i - \mathbf{b}_i$, thus obtaining:

$$(\mathbf{c}_1 - \mathbf{b}_1)^{\mathrm{T}} \, \dot{\mathbf{p}} = (\mathbf{c}_1 - \mathbf{b}_1)^{\mathrm{T}} \frac{\mathbf{b}_1 - \mathbf{a}_1}{\|\mathbf{b}_1 - \mathbf{a}_1\|} \dot{\rho}_1 \tag{2a}$$

$$(\mathbf{c}_2 - \mathbf{b}_2)^{\mathrm{T}} \, \dot{\mathbf{p}} = (\mathbf{c}_2 - \mathbf{b}_2)^{\mathrm{T}} \frac{\mathbf{b}_2 - \mathbf{a}_2}{\|\mathbf{b}_2 - \mathbf{a}_2\|} \dot{\rho}_2 \tag{2b}$$

$$(\mathbf{c}_3 - \mathbf{b}_3)^{\mathrm{T}} \, \dot{\mathbf{p}} = (\mathbf{c}_3 - \mathbf{b}_3)^{\mathrm{T}} \frac{\mathbf{b}_3 - \mathbf{a}_3}{\|\mathbf{b}_3 - \mathbf{a}_3\|} \dot{\rho}_3 \tag{2c}$$

Equations (2a), (2b) and (2c) can now be cast in vector form, namely,

$$\mathbf{A} \, \dot{\mathbf{p}} = \mathbf{B} \, \dot{\mathbf{p}}$$

where $\mathbf{A}$ and $\mathbf{B}$ are the parallel and serial Jacobian matrices, respectively:

$$\mathbf{A} = \begin{bmatrix} (\mathbf{c}_1 - \mathbf{b}_1)^{\mathrm{T}} \\ (\mathbf{c}_2 - \mathbf{b}_2)^{\mathrm{T}} \\ (\mathbf{c}_3 - \mathbf{b}_3)^{\mathrm{T}} \end{bmatrix} \text{ and } \mathbf{B} = \begin{bmatrix} (\mathbf{c}_1 - \mathbf{b}_1)^{\mathrm{T}} \frac{\mathbf{b}_1 - \mathbf{a}_1}{\|\mathbf{b}_1 - \mathbf{a}_1\|} & 0 & 0 \\ 0 & (\mathbf{c}_2 - \mathbf{b}_2)^{\mathrm{T}} \frac{\mathbf{b}_2 - \mathbf{a}_2}{\|\mathbf{b}_2 - \mathbf{a}_2\|} & 0 \\ 0 & 0 & (\mathbf{c}_3 - \mathbf{b}_3)^{\mathrm{T}} \frac{\mathbf{b}_3 - \mathbf{a}_3}{\|\mathbf{b}_3 - \mathbf{a}_3\|} \end{bmatrix} \tag{3}$$

When $\mathbf{A}$ and $\mathbf{B}$ are not singular, we can also study the Jacobian kinematic matrix $\mathbf{J}$ (Merlet 1997) to optimise the manipulator,

$$\dot{\mathbf{p}} = \mathbf{J} \, \dot{\mathbf{p}} \text{ with } \mathbf{J} = \mathbf{A}^{-1} \, \mathbf{B} \tag{4a}$$

or the inverse Jacobian kinematic matrix $\mathbf{J}^{-1}$, such that

$$\dot{\boldsymbol{\rho}} = \mathbf{J}^{-1}\,\dot{\mathbf{p}} \text{ with } \mathbf{J}^{-1} = \mathbf{B}^{-1}\,\mathbf{A} \tag{4b}$$

The parallel singularities (Chablat and Wenger 1998) occur when the determinant of the matrix $\mathbf{A}$ vanishes, i.e. when $\det(\mathbf{A})=0$. In such configurations, it is possible to move locally the mobile platform whereas the actuated joints are locked. These singularities are particularly undesirable, because the structure cannot resist any force and control is lost. From eq. (3), it is apparent that the parallel singularities occur whenever the points $C_i$ and $B_i$ are coplanar:

$$(\mathbf{c}_1 - \mathbf{b}_1) = \alpha\,(\mathbf{c}_2 - \mathbf{b}_2) + \lambda\,(\mathbf{c}_3 - \mathbf{b}_3) \tag{5}$$

or when the links $B_i C_i$ are parallel:

$$(\mathbf{c}_1 - \mathbf{b}_1)\ /\!/\ (\mathbf{c}_2 - \mathbf{b}_2) \text{ and } (\mathbf{c}_2 - \mathbf{b}_2)\ /\!/\ (\mathbf{c}_3 - \mathbf{b}_3) \text{ and } (\mathbf{c}_3 - \mathbf{b}_3)\ /\!/\ (\mathbf{c}_1 - \mathbf{b}_1). \tag{6}$$

These configurations cannot be reached with the design of the manipulator studied.

Serial singularities arise when the serial Jacobian matrix, $\mathbf{B}$, is no longer invertible, i.e., when $\det(\mathbf{B})=0$. At a serial singularity, a direction exists along which any cartesian velocity cannot be produced. From equation (3), it is apparent that $\det(\mathbf{B})=0$ when, for one leg i, $(\mathbf{b}_i - \mathbf{a}_i) \perp (\mathbf{c}_i - \mathbf{b}_i)$. It is possible to avoid such singularities by adjusting the joint limits of the linear actuated joints.

## 3    Performance analysis of the orthoglide

### 3.1    Notion of Condition number

The *condition number* of an $m \times n$ matrix $\mathbf{M}$ with $m \le n$, $\kappa(\mathbf{M})$ can be defined in various ways; for our purposes, we define $\kappa(\mathbf{M})$ as the ratio of the largest, $\sigma_L$, to the smallest, $\sigma_S$ singular values of $\mathbf{M}$ (Golub 89),

$$\kappa(\mathbf{M}) = \sqrt{\frac{\sigma_L}{\sigma_S}} \tag{7}$$

The singular values $\{\sigma_k\}_1^m$ of matrix $\mathbf{M}$ are defined as the square roots of the nonnegative eigenvalues of the positive semi-definite $m \times m$ matrix $\mathbf{M}\,\mathbf{M}^T$.

For the purpose of design and performances analysis, we need to define the condition number of the Jacobian matrix. The condition number of the Jacobian matrix is an interesting performance index since it characterises the distortion of a unit ball under the transformation represented by the Jacobian matrix at hand (Angeles 1997). The Jacobian matrix is said to be isotropic when its condition number attains its minimum value of one (there is no distortion). We know that the Jacobian matrix of a manipulator is used to relate (i) the joint rates and the Cartesian velocities, and (ii) the static load on the output link and the joint torques or forces. Thus, the condition number of the Jacobian matrix can be used to measure the uniformity of the distribution of the tool velocities and forces in the Cartesian workspace.

### 3.2    Condition number of the inverse Kinematic Jacobian Matrix

For parallel manipulators, it is more convenient to study the conditioning of the Jacobian matrix that is related to the inverse transformation, which we have called inverse kinematic Jacobian matrix $\mathbf{J}^{-1}$ in equation (4b). In our case, matrices $\mathbf{B}$ and $\mathbf{J}^{-1}$ can be derived easily as follows:

$$\mathbf{B}^{-1} = \begin{bmatrix} \dfrac{1}{\eta_1} & 0 & 0 \\ 0 & \dfrac{1}{\eta_2} & 0 \\ 0 & 0 & \dfrac{1}{\eta_3} \end{bmatrix} \text{ with } \eta_i = (\mathbf{c}_i \text{-} \mathbf{b}_i)^T \dfrac{\mathbf{b}_i \text{-} \mathbf{a}_i}{\|\mathbf{b}_i \text{-} \mathbf{a}_i\|} \text{ and } \mathbf{J}^{-1} = \begin{bmatrix} \dfrac{1}{\eta_1} (\mathbf{c}_1 \text{-} \mathbf{b}_1)^T \\ \dfrac{1}{\eta_2} (\mathbf{c}_2 \text{-} \mathbf{b}_2)^T \\ \dfrac{1}{\eta_3} (\mathbf{c}_3 \text{-} \mathbf{b}_3)^T \end{bmatrix} \qquad (5)$$

The matrix $\mathbf{J}^{-1}$ is isotropic when:

$$\frac{1}{\eta_1} \|\mathbf{c}_1 \text{-} \mathbf{b}_1\| = \frac{1}{\eta_2} \|\mathbf{c}_2 \text{-} \mathbf{b}_2\| = \frac{1}{\eta_3} \|\mathbf{c}_3 \text{-} \mathbf{b}_3\| \qquad (6a)$$

$$(\mathbf{c}_1 \text{-} \mathbf{b}_1)^T (\mathbf{c}_2 \text{-} \mathbf{b}_2) = 0, \ (\mathbf{c}_2 \text{-} \mathbf{b}_2)^T (\mathbf{c}_3 \text{-} \mathbf{b}_3) = 0, \ (\mathbf{c}_3 \text{-} \mathbf{b}_3)^T (\mathbf{c}_1 \text{-} \mathbf{b}_1) = 0 \qquad (6b)$$

Equation (6a) states that the orientation between the axis of the prismatic joint and the link $B_iC_i$ must be the same for each leg i. Equation (6b) means that the links $B_iC_i$ must be orthogonal to one another.

We note that there is no condition on the lengths of the link $B_iC_i$. We set them to be equal to one another, in order to have a symmetric design. On the other hand, the orthogonal orientation of the relative prismatic joints cannot be deduced from the isotropic condition. It can be defined by the manipulability analysis, as shown in the following section.

### 3.3    Manipulability analysis

In the case of serial PPP machine tool, a motion of an actuated joint yields the same motion of the tool. For a parallel machine, these motions are not equivalent. When the machine is close to a parallel singularity, a small joint rate can generate a large velocity of the tool. This means that the positioning accuracy of the tool is lower in certain directions near parallel singularities because the encoder resolution is amplified. In addition, a velocity amplification in one direction is equivalent to a loose of rigidity in this direction. The manipulability ellipsoids of the Jacobian matrix of robotic manipulators was defined several years ago by (Salisbury and Craig 1982) and (Yoshikawa 1985). This concept has then been applied as a performance index to parallel manipulators (Kim et al 1997). Note that, although the concept of manipulability is close to the concept of condition number, they do not provide the same information. The condition number quantifies the proximity to an isotropic configuration, i.e. where the ellipsoid is a sphere, or, in other words, where the velocity amplification is equal in any direction, but it does not provide information as to the magnitude of the velocity amplification. The manipulability ellipsoid of $\mathbf{J}^{-1}$ will be used here for (i) justifying the orthogonal orientation of the prismatic joints and (ii) defining the joint limits of the ortholide such that the maximal velocity amplification remains under a reasonable limit. We want that the velocity amplification factor and the force amplification factor be equal to one at the isotropic configuration. This condition implies that the three terms of equation (6a) must be equal to one:

$$\frac{1}{\eta_1} \|\mathbf{c}_1 \text{-} \mathbf{b}_1\| = \frac{1}{\eta_2} \|\mathbf{c}_2 \text{-} \mathbf{b}_2\| = \frac{1}{\eta_3} \|\mathbf{c}_3 \text{-} \mathbf{b}_3\| = 1 \qquad (7)$$

which implies that $(\mathbf{b}_i \text{-} \mathbf{a}_i)$ and $(\mathbf{c}_i \text{-} \mathbf{b}_i)$ must be collinear for each i. Since, at the isotropic configuration, links $B_iC_i$ are orthogonal to one another, (7) implies that the links $A_iB_i$ are orthogonal, i.e. the prismatic joints are orthogonal.

By using equation (4b), we can write a relation between the velocity $\dot{\mathbf{p}}$ of point $P$ and the joint rates $\dot{\tilde{\mathbf{p}}}$. For joint rates belonging to a unit ball, namely, $\dot{\tilde{\mathbf{p}}} \leq 1$, the Cartesian velocities belong to an ellipsoid such that:

$$\dot{\mathbf{p}}^T (\mathbf{J} \mathbf{J}^T)^{-1} \dot{\mathbf{p}} \leq 1 \qquad\qquad (8)$$

The eigenvectors of matrix $(\mathbf{J} \mathbf{J}^T)^{-1}$ define the direction of its principal axes and the square roots $\xi_1$, $\xi_2$ and $\xi_3$ of the eigenvalues of $(\mathbf{J} \mathbf{J}^T)^{-1}$ are the lengths of the aforementioned principal axes. The factors of velocity amplification in the directions of the principal axes are defined by $\psi_1 = 1 / \xi_1$, $\psi_2 = 1 / \xi_2$ and $\psi_3 = 1 / \xi_3$

To limit the variations of this factor in the Cartesian workspace, we impose $1/3 \leq \psi_i \leq 3$ all over the workspace. This condition leads to the definition of joint limits for the prismatic joints (Chablat et al 2000).

### 3.4    Cartesian Workspace analysis

The Cartesian workspace is one of the most important performance evaluation criteria of PKM. However, the Cartesian workspace definition commonly considered (set of the reachable configurations of the output link) is insufficient to asses the real performances of a PKM since this definition does not take into account the possibility to execute motions inside the Cartesian workspace. This is of primary importance for parallel mechanisms which generally feature internal singularities which should not be crossed. Self collisions may also arise. To cope with the moveability of a manipulator, the connected regions of the Cartesian workspace were defined for serial manipulators (Wenger and Chedmail 1991) and, more recently, for parallel manipulators (Chablat and Wenger 1998). The Cartesian workspace connectivity can be further analysed according to which type of motions should be studied: the n-connectivity is intended to point-to-point motions and the t-connectivity is suitable for continuous trajectory tracking, like in arc-welding or machining. In the context of machine tool, the t-connected regions must be considered. More precisely, the t-connected regions of a PKM are the regions of the Cartesian workspace which are free of singularity (and collisions) and where any continuous path is feasible. The size and shape of the maximal t-connected region is of primary importance for the global geometric performances evaluation of a machine tool. The orthoglide has been designed such that its Cartesian workspace is free of singularities and self-collisions. Thus, the maximal t-connected region of the orthoglide is its Cartesian workspace. Figure 3 shows the Cartesian workspace of the orthoglide, and figure 4 depicts a cross section. The Cartesian workspace has a fairly regular shape which is close to a cube. We have used here our general octree-based algorithm for the calculation of the workspace (Chablat 1998). If there is no obstacle to take into account, the workspace of the orthoglide can be easily calculated analytically by its boundaries.

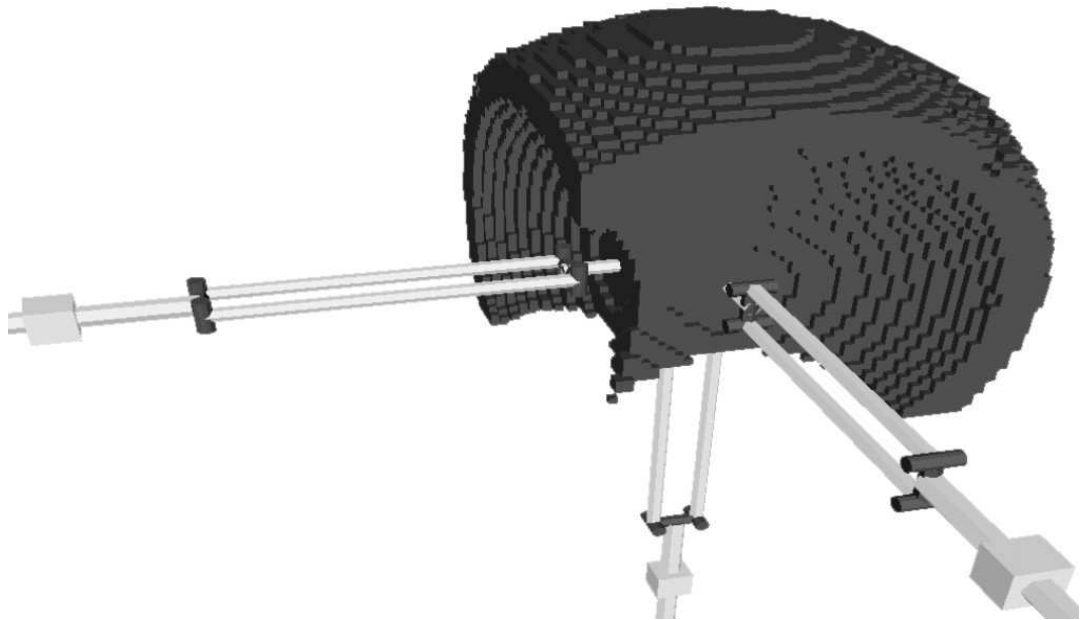

*Figure 3: Cartesian workspace of Orthoglide manipulator using an octree model*

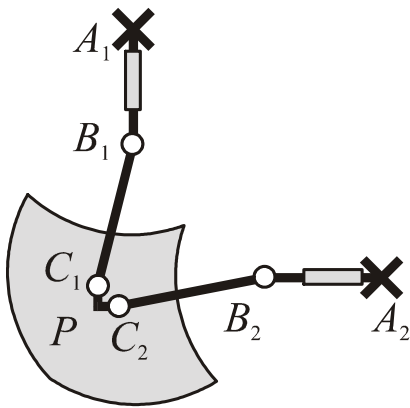

*Figure 4: Cross section of the Cartesian workspace*

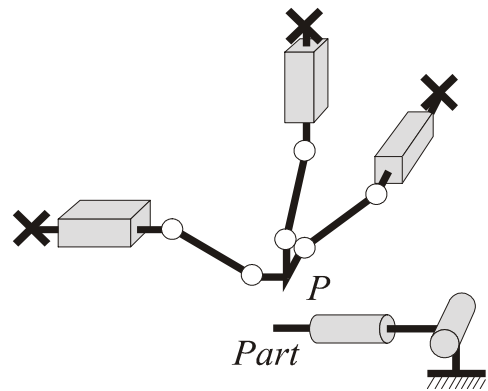

*Figure 5: Extension to 5-axis machining*

## 4 Extension to 5-axis machines

The orthoglide described above is dedicated to 3-axis machining applications. The machining of geometrically complex parts like moulds requires 5-axis machines. The orthoglide can be extended to a 5-axis machine by adding a 2-axis orienting branch to the initial positioning structure. This can be done in two ways. The first approach consists in mounting this branch serially at the moving body. We get a hybrid kinematic structure, like the Tricept 805 (see §**Erreur ! Source du renvoi introuvable.**). However, this solution increases the moving masses since the orienting device must be carried by the positioning structure. Another solution is to mount the orienting branch on the base of the orthoglide, as shown in figure 5.

# 5 Conclusions

Presented in this paper is new kinematic structure of PKM dedicated to machining applications: the Orthoglide. The main feature of this PKM is its compromised design between the popular serial PPP architecture and the parallel kinematic arrangement of the links. The workspace is simple and regular and features no singularity nor self-collisions. The orthoglide has been designed such that its Jacobian matrix is isotropic in the centre of its workspace. Most existing PKM suffer from high variations of speed and rigidity performances in their workspace (Wenger et al 1999). The velocity amplification factor of the orthoglide is one at the isotropic point and bounded by reasonable values in the rest of the workspace. Further stiffness and sensitivity analyses will be conducted by the authors, taking into account the kinematics of the parallelograms.

# References

Stewart D., 1965, "A Platform with 6 Degrees of Freedom", *Proc. of the Institution of Mechanical Engineers*, 180(Part 1, 15), pp. 371-386.

Gosselin C. and Wang J., 1998, "On the Design of Gravity-Compensated Six-Degree-of-Freedom Parallel Mechanisms", *Int. IEEE Int. Conf. on Robotics and Automation*, pp. 2287-2294.

Clavel, 1988, "DELTA, a Fast Robot with Parallel Geometry", *Pro. Of the 18th Int. Symp. of Robotic Manipulators*, IFR Publication, pp. 91-100.

Wenger P., Gosselin C. and Maille B., 1999, "A Comparative Study of Serial and Parallel Mechanism Topologies for Machine Tools", *Proc. PKM'99,* Milano, 1999, pp 23-32.

Merlet J-P., 1997, *Les robots parallèles*, 2$^{nd}$ édition, Hermes, Paris, 1997.

Chablat D., Wenger P. and Angeles J., 2000, "Conception Isotropique d'une morphologie parallèle: Application à l'usinage*", 3$^{th}$ Int. Conf. On Integrated Design and Manufacturing in Mechanical Engineering*, Montreal, May 2000.

Chablat D. and Wenger P., 1998, "Working Modes and Aspects in Fully-Parallel Manipulator", *Pro. IEEE Int. Conf. On Robotics and Automation*, pp. 1964-1969.

Golub, G-H. and Van Loan, C-F., 1989, *Matrix Computations*, The John Hopkins University Press, Baltimore.

Angeles J., 1997, *Fundamentals of Robotic Mechanical Systems*, Springer-Verlag, New York.

Salisbury J-K. and Craig J-J., 1982, "Articulated Hands: Force Control and Kinematic Issues", *The Int. J. Robotics Res.*, Vol. 1, No. 1, pp. 4-17.

Yoshikawa T., 1985, "Manipulability of Robotic Mechanisms", *The Int. J. Robotics Res.*, Vol. 4, No. 2, pp. 3-9.

Kim J., Park C., Kim J. and Park F.C., 1997,"Performance Analysis of Parallel Manipulator Architectures for CNC Machining Applications", *Proc. IMECE Symp. On Machine Tools,* Dallas.

Wenger P. an Chedmail P., 1991, "Ability of a Robot to Travel through its Free Workspace", *The Int. J. of Robotic Research*, Vol. 3, No 10, pp. 214-227.